\documentclass[sigconf]{acmart}

\AtBeginDocument{%
  \providecommand\BibTeX{{%
    \normalfont B\kern-0.5em{\scshape i\kern-0.25em b}\kern-0.8em\TeX}}}

\acmConference[Conference acronym 'XX]{}{}{}
\usepackage{float}
\usepackage{makecell}
\usepackage{algorithm} 
\usepackage{url}
\usepackage{algpseudocode} 
 \usepackage[stable]{footmisc}
\usepackage{amsfonts}       
\usepackage{nicefrac}       
\usepackage{microtype}      
\usepackage{graphicx,amsmath,amsfonts,amsthm,amscd,bm,url,color,latexsym}
\usepackage{caption}
\usepackage{subcaption}
\usepackage{multirow}
\usepackage[capitalize,nameinlink]{cleveref}

\newcommand{\mb}{\mathbf}

\begin{document}

\title{Genetics-Driven Personalized Disease Progression Model}






\author{Haoyu Yang}
\affiliation{%
  \institution{University of Minnesota}
  \country{USA}
  }
\email{yang6993@umn.edu}

\author{Sanjoy Dey}
\affiliation{%
  \institution{IBM Research}
  \country{USA}
  }
\email{deysa@us.ibm.com}

\author{Pablo Meyer}
\affiliation{%
  \institution{IBM Research}
  \country{USA}
  }
\email{pmeyerr@us.ibm.com}



\begin{abstract}
Modeling disease progression through multiple stages is critical for clinical decision-making for chronic diseases, e.g., cancer, diabetes, chronic kidney diseases, and so on. Existing approaches often model the disease progression as a uniform trajectory pattern at the population level. However, chronic diseases are highly heterogeneous and often have multiple progression patterns depending on a patient’s individual genetics and environmental effects due to lifestyles. We propose a personalized disease progression model to jointly learn the heterogeneous progression patterns and groups of genetic profiles. In particular, an end-to-end pipeline is designed to simultaneously infer the characteristics of patients from genetic markers using a variational autoencoder and how it drives the disease progressions using an RNN-based state-space model based on clinical observations. Our proposed model shows improvement on real-world and synthetic clinical data.
\end{abstract}






\maketitle

\section{Introduction}
Chronic diseases such as Diabetes, Cancer and Huntington’s disease typically progress through multiple stages, ranging from mild to moderate to severe. A better understanding of progression of chronic diseases is crucial at multiple stages of clinical decision making such as early detection, preventive interventions, and precision medicine. Modeling the disease progression longitudinally from large scale patient records can also unravel important subtypes of the disease by representing its impact on sub-populations, from the rate of progression. Such disease progression models (DPM) have been used in clinical decision making for multiple chronic diseases such as Alzheimer's disease \cite{sukkar2012disease, liang2021rethinking, el2020multimodal}, Huntington's Disease \cite{sun2019probabilistic, mohan2022machine} and prostate cancer \cite{lee2022developing}. Also, DPM has been used to assess the impact of novel therapeutics on the course of disease, thus elucidating a potential clinical benefit in the drug discovery process \cite{barrett2022role, goteti2023opportunities} and in designing clinical trials \cite{miller2005modeling}.  

 A few recent studies used deep learning based state-space models for DPMs, where multi-dimensional time-varying representations were used to learn disease states \cite{krishnan2017structured}. For example, an attentive state-space model based on two recurrent neural networks (RNN) was used to remember the entire patients medical history from past electronic health records (EHR) which will ultimately govern transitions between states \cite{alaa2019attentive}. Similarly, the representations of disease states were conditioned on patients treatment history to analyze the impact of diverse pharmaco-dynamic effects of multiple drugs \cite{hussain2021neural}. In particular, they used multiple attention schemes, each modeling a pharmaco-dynamic mechanism of drugs, which ultimately determined the state representations of DPMs. 

However, one caveat with these existing DPMs is their assumption that disease trajectory patterns are uniform across all patients as they progress through multiple stages.
In reality, progression patterns differ widely from patient to patient due to their inherent genetic predisposition and environmental effects derived from to patient's lifestyle \cite{dey2013integration}. 
Previous approaches for disease progression pathways are often not personalized, rather they model disease progression as a uniform trajectory pattern at the population level. 
Although some recent proposed models, such as \cite{hussain2021neural}, conditioned DPM on other available clinical observations including genetic data, demographics and treatment patterns, these models did not aim at building personalized DPMs based on patients genetic data. 

In this work,  our objective is to build a genetics-driven personalized disease progression model (PerDPM), where the progression model is built separately for each patient's group associated with different inherent genetic makeups. 
The inherent characteristics of patients that drive the disease progression model differently are defined by their ancestry and inherent genetic composition. 
Specifically, we developed a joint learning framework, where the patient's genetic grouping is discovered from large-scale genome-wide association studies (GWAS), along with developing disease progression models tailored to each identified genetic group. 
To achieve this, we first use a variational auto-encoder which identifies different groups of genetic clusters from GWAS data. Then, we model the disease progression model as a state-space model where the state transitions are dependent upon the genetic makeups, treatments and clinical history available so far. Our proposed model is generic enough to discover the genetic groupings and their associated progression pathway from clinical observations by utilizing both static (e.g., genetic data) and longitudinal healthcare records such as treatments, diagnostic variables measuring co-morbidities, and any clinical assessment of diseases such as laboratory measurements. 

The main contributions of this paper can be summarized as below: 
\begin{itemize}
    \item We proposed a genetic-driven disease personalized progression model (PerDPM). In PerDPM, we designed two novel modules: one for genetic makeups inference and another for genetics driven state transition modeling. 
    \item We introduced a joint learning optimization framework to infer both genetic makeups and latent disease states. This framework simultaneously performs clustering of genetic data and develops the genetic dependent disease progression model.

    \item The proposed framework was tested both in synthetic and one of largest real-world EHR cohort collected from the UK called UKBioBank (UKBB). The effectiveness of the model was tested on this UKBB dataset in the context of modeling Chronic Kidney Disease.
\end{itemize}

\section{Related Work}
Disease progression models (DPM) provide a basis for learning from prior clinical experience and summarizing knowledge in a quantitative fashion. 
We refer to review article \cite{hainke2011disease} for general overview of DPM on multiple domains. The line of research for building DPM from longitudinal records belongs to the machine learning and artificial intelligence, where the state transitions are learned in a probabilistic manner. The earliest DPM technique was built upon Hidden Markov Models \cite{di2007stochastic} to infer disease states and to estimate the transitional probabilities between them simultaneously. These baseline DPMs are mostly applicable for regular time intervals, while clinical observations are often collected at irregular intervals. A continuous-time progression model from discrete-time observations to learn the full progression trajectory has been proposed for this purpose \cite{wang2014unsupervised, Bartolomeo2011}. In \cite{liu2013longitudinal} a 2D continuous-time Hidden Markov Model for glaucoma progression modeling from longitudinal structural and functional measurements is proposed. However, all of these probabilistic generative models are computationally expensive and not applicable for large-scale data analysis. 

\begin{figure}[h]
    \centering
    \includegraphics[width = 0.5\linewidth]{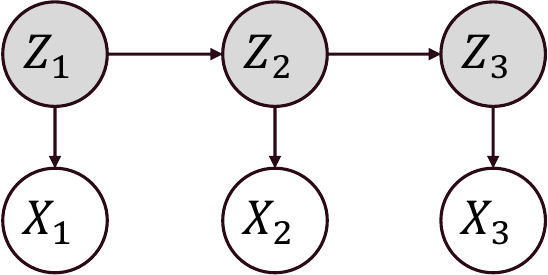}
    \caption{State-Space Model (SSM). Observations $X_t$ are generated from latent variables $Z_t$. Lines denote the generative process. Different from RNNs structure, the latent representation $\mb Z_t$ in SSM is not deterministic.}
    \label{fig:ssm}
\end{figure}

Recently, deep learning based generative Markov model such as state-space model (SSM) has been proposed for learning DPM from large-scale data.
As shown in \cref{fig:ssm}, the observational variables are generated from latent variables through emission models, and the transitions between latent variables correspond to the underlying dynamics of the system. 
Recent work added deep neural networks structure to linear Gaussian SSM to learn non-linear relationships from high dimensional time-series \cite{krishnan2017structured, rangapuram2018deep, ansari2021deep}. 

Interpretability has made SSM popular in modeling the progression of a variety of chronic diseases. For different diseases and different tasks, the design of SSM varies. In \cite{alaa2019attentive}, a transition matrix in their model is used to learn interpretable state dynamics, as \cite{severson2020personalized} introduces patient specific latent variables to learn personalized medication effects and \cite{hussain2021neural, qian2021integrating} focuses on pharmacodynamic model and integrating expert knowledge. To the best of our knowledge, none of these techniques is able to model how genetic heterogeneity may impact the disease progression directly through the integration of large-scale clinical and genomic data.

\section{Notations}
\begin{table}[]
\caption{Notations lookup table}
    \centering
    \begin{tabular}{|c|c|c|}
    \hline
         Variables &Dimensions& Note  \\\hline
         $\mb X$ & $N \times T\times  d_x$& Temporal clinical features\\\hline
         $\mb U$ & $N \times T\times  d_u$& Temporal treatments\\\hline
         $\mb Z$ & $N \times T\times  d_z$& Temporal latent states\\\hline
         $\mb G$ & $N \times d_g$ & Genetic features\\\hline
         $\mb V$ & $N \times K$& Genetic group assignment matrix\\\hline
         $\mb Y$ & $-$& Set of observable variables $\{\mb X, \mb G, \mb U\}$\\\hline
    \end{tabular}
    
    \label{tab:lookup}
\end{table}

Let $x_i \in \mathbb{R}^{d_x}$ denote all the clinical features for a particular patient i, where $i \in \{1,2,...,N\}$, $d_x$ and $N$ represent the total number of clinical features and samples, respectively. Let $X = [x_1; x_2; ..., x_N] \in \mathbb{R}^{N \times d_x}$ denote the whole matrix of clinical observations from EHR. Also, EHR data are typically observed at multiple irregular time points. So, we use $X_{i,t}$ to denote the observed clinical features at time point t for the $i$-th patient, where $t \in \{1,2,...,T_i\}$ and $T_i$ denotes the total number of visits made by the patient $i$. 
Through the paper, we will omit the sample index $i$ when there is no ambiguity.
These clinical observations can consist of any temporal measurements such as prior disease history, lab assessments for the particular disease being model, image scans, etc. Let $\mb X\in\mathbb{R}^{N\times T\times d_x}$ denote the final matrix of clinical observation containing all temporal data, where $T$ is the largest number of visits of patients. Similarly, $\mb U\in\mathbb{R}^{N\times T\times d_u}$ denotes all treatments collected temporarily, where $d_u$ is the dimensions of all possible treatments performed on the patients. 
$\mb G \in \mathbb{R}^{N\times d_g}$ denotes genetic matrix, where $d_g$ is the total number of genetic features observed for patients. Note that genetics features are observed only during the initial visit of the patient, so genetic matrix $\mb G$ don't have any temporal dimensions associated with them. Further, we assume the whole population can be divided into $K$ different groups, expressing different disease progression dynamics and implicitly encoded by genetic information. In our model. we use a proxy variable $\mb V\in\mathbb{R}^{N\times K}$ to represent the probability of group assignment for each patient.
Give patient's clinical features $\mb X$, treatments $\mb U$, and genetic information $\mb G$, the final task is to learn latent representations $\mb Z\in\mathbb{R}^{N\times T\times d_z}$, which describe the hidden states of the disease progression.
For better understanding, we provide a lookup table for these notation in \cref{tab:lookup}.

\begin{figure*}[ht!]
    \centering
    \includegraphics[width=0.8\textwidth]{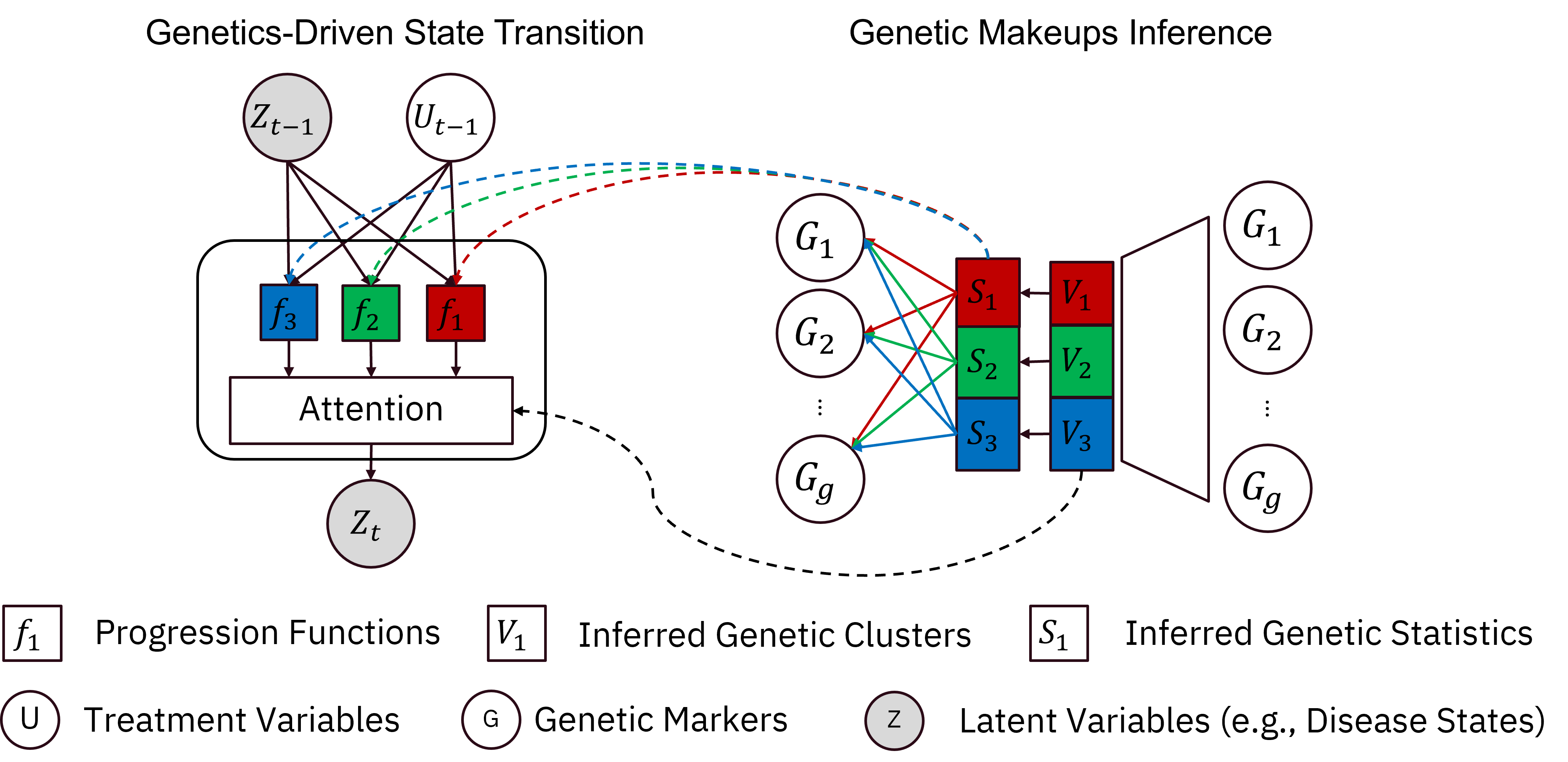}
    \caption{Architecture of the proposed model. The model has two key components: genetic makeups inference and genetics-driven state transition. The other components, e.g., inference network and emission model, are omitted in this figure.}
    \label{fig:overview}
\end{figure*}

\section{Method}
In this study, we want to extend the above mentioned state-space based disease progression models to a genetics driven personalized DPM model. Specifically, we want to make sure that the transitions between state $\mb Z$s are conditioned on the genetic profile that is leaned from GWAS data as shown in \cref{fig:overview}. In addition to that, we also assume that the transition between states are dependent on the treatment patterns similar to \cite{hussain2021neural}, so that the disease progression model can be better characterized by the genetics makeup of the patient. 

Our proposed framework first models the genetic groupings from the available high-dimensional GWAS data using a variational auto-encoder, and then estimates the representations of disease progression through a generative framework with the above-mentioned assumptions of progression being conditioned on these genetic makeups and treatments. 

In the rest of this section, we will first introduce our framework by factorizing the joint distribution of all observable variables according to previously mentioned methods. 
Then, we will derive the evidence lower bound (ELBO) for variational inference. 
Lastly, we will provide details on the generative model and the inference model for estimating genetic makeups and the disease states transitions. 

\subsection{Overall Framework}
To characterize the disease progression using state-space models, a patient is modeled to be in a disease state $\mb Z_t$ at each time step $t$, which is manifested in clinical observation $\mb X_t$. 
Specifically, in terms of observational data, we assume the clinical observations $\mb X$ are generated from latent variables $\mb Z$ through an \textit{emission model}. The disease progression from $\mb Z_{t-1}$ to $\mb Z_{t}$ is governed by a \textit{transition model} and some observable variables such as treatments $\mb U$ and genetic markers $\mb G$. 
Throughout the paper, we assume that the states are hidden and thus will be inferred in an unsupervised fashion.

We first model the joint distribution of all observable variables via the following factorization:
\begin{align}
    \label{eq:factorization_step1}
    p(\mb X, \mb G, \mb U) = \Pi_{t=0}^{T}p(\mb X_t, \mb G| \mb U_t) p(\mb U_t)
\end{align}

To further factorize the joint distribution $p(\mb X_t, \mb G | \mb U_t)$, we introduce two latent variables: latent states $\mb Z$ and genetic cluster variable $\mb V$, to mimic the real data generation process. The variable $\mb V \in \mathbb{R}^{N \times K}$ describes the likelihood of a patient belonging to a certain genetic group.
Then, term $p(\mb X_t, \mb G | \mb U_t)$ can be further factorized as follow:

\begin{align}
    \label{eq:generative3}
    &p(\mb X_t, \mb G |\mb U_t) \\\nonumber
    =&\int_{\mb Z, \mb V} p(\mb X_t, \mb G |\mb Z_t, \mb U_t, \mb V)p(\mb Z_t, \mb V |\mb Z_{t-1},  \mb U_t)d\mb Z d\mb V \\\nonumber
    =&\int_{\mb Z, \mb V}  \{p(\mb X_t|\mb Z_t, \mb U_t, \mb V)p(\mb G |\mb Z_t, \mb U_t, \mb V)p(\mb Z_t |\mb Z_{t-1}, \mb U_t, \mb V) \\ \nonumber
    &  p(\mb V| \mb U_t)\}d\mb Z d\mb V \\\nonumber
    =&\int_{\mb Z, \mb V}  p(\mb X_t| \mb Z_t)p(\mb G | \mb V)p(\mb Z_t |\mb Z_{t-1}, \mb U_t, \mb V)p(\mb V)d\mb Z d\mb V
\end{align}

Here, the first equation follows our assumption that the joint distribution of $\mb Z$ and $\mb V$ is conditioned the previous latent state and current treatment, which is the common assumption made by state space models.
Then, in the second equation, we assumed the observations $\mb X$ is conditionally-independent of genetic information $\mb G$ given latent variables $\mb V, \mb Z$ and treatment $\mb U$.
Similarly, in the last equation, we assumed the observations $\mb X$ is conditionally-independent of all other variables given latent state variables $\mb Z$, the genetic information $\mb G$ is conditionally-independent of all other variables given the latent proxy variable $\mb V$, and the latent proxy variable $\mb V$ is independent of treatment $\mb U$

By combining \cref{eq:factorization_step1} and \cref{eq:generative3}, we obtain the joint probability of all the observable variables is
\begin{align}
\label{eq:likelihood_overall}
    & p(\mb X, \mb G, \mb U) = \Pi_{t=0}^{T}p(\mb X_t, \mb G| \mb U_t) p(\mb U_t)\nonumber\\
    =&\int_{\mb Z, \mb V}\Pi_{t=0}^Tp(\mb U_t)  p(\mb X_t| \mb Z_t)p(\mb G | \mb V)
    p(\mb Z_t |\mb Z_{t-1}, \mb U_t, \mb V)p(\mb V)d\mb Z d\mb V 
\end{align}

To make sure the edge case makes sense, we use a special setting for the initial latent state, i.e., $p(\mb Z_0 |\mb U_t, \mb V)$.

\subsection{Derivation of Evidence Lower Bound (ELBO)}
Directly maximizing the likelihood shown in \cref{eq:likelihood_overall} is intractable. Instead, we learn via maximizing a variational lower bound (ELBO).
For simplicity, let $\mb Y$ represents all observable variables, i.e., $\mb Y = \{\mb X, \mb G, \mb U\}$, and $\mathcal D$ represent the dataset, the ELBO is 
\begin{align}
\label{eq:likelihood_and_elbo}
    \log p(\mb Y) &\geq ELBO\nonumber\\
    &=\mathbb{E}_{\mb Y\sim\mathcal{D}, q_\phi(\mb Z, \mb V|\mb Y)}[\log p(\mb Y|\mb Z, \mb V)] \nonumber\\
    &- 
     \mathbb{E}_{\mb Y\sim\mathcal{D}}[KL(q_\phi(\mb Z, \mb V|\mb Y)||p_\theta(\mb Z, \mb V))]
\end{align}
where $q_\phi$ and $p_\theta$ are learned posterior and prior distributions.
Using  \cref{eq:likelihood_overall} to expand \cref{eq:likelihood_and_elbo} yields
\begin{align}
    ELBO =& \mathbb{E}_{\mb Y\sim\mathcal{D}, q_\phi(\mb Z, \mb V|\mb Y)}[\sum_t(\log p(\mb X_t| \mb Z_t) + \log p(\mb G | \mb V))] \nonumber\\
    &- \mathbb{E}_{\mb Y\sim\mathcal{D}}[\sum_t KL(q_\phi(\mb Z_t, \mb V|\mb Y_t, \mb Z_{t-1})||p_\theta(\mb Z_t, \mb V))]\nonumber
\end{align}

We further factorize the ELBO into two components as follow:

\begin{align}
\centering
\label{eq:elbo_final}
    &\quad ELBO = \underbrace{ T\mathbb{E}_{\mb Y\sim\mathcal{D}, q_\phi(\mb Z, \mb V|\mb Y)}[\log p_\theta(\mb G | \mb V)]  }_{\text{Modeled by VAE}}\nonumber\\
    &- \underbrace{  T\mathbb{E}_{\mb Y\sim\mathcal{D}}[KL(q_\phi(\mb V|\mb G)|| p_\theta(\mb V))] }_{\text{Modeled by VAE}}\nonumber\\
     &+ \quad \underbrace{\mathbb{E}_{\mb Y\sim\mathcal{D}, q_\phi(\mb Z, \mb V|\mb Y)}[\sum_t\log p_\theta(\mb X_t| \mb Z_t)]}_{\text{Modeled by State Space Model}}\nonumber\\
    &-\quad \underbrace{\sum_t \mathbb{E}_{\mb Y\sim\mathcal{D}, q_\phi(\mb V|\mb G)}[KL( q_\phi(\mb Z_t|\mb Y_t, \mb Z_{t-1})|| p_\theta(\mb Z_t|\mb Z_{t-1}, \mb V))] }_{\text{Modeled by State Space Model}}
\end{align}

As shown in the equation, we parameterized these two components using a VAE structure and state space models. The detailed derivation is given in the Supplementary section.

To minimize the ELBO, we follow the structural inference method proposed by \cite{krishnan2017structured}. The key idea is to parameterize the inference models $q_\phi$ and generative models $p_\theta$ using differentiable neural network and update $\phi, \theta$ jointly using stochastic gradient descent. For the rest of this section, we introduce our design for inference models $q_\phi$ and generative models $p_\theta$.

\subsection{Genetic Makeups Inference}
Typically, genetic data is collected in genome-wide association studies (GWAS), which measure the genetic variation of a person's genome at a single base position. These variations, termed Single Nucleotide Polymorphisms (SNPs), impact how and to what degree a particular trait or disease phenotype is manifested in an individual. One particular challenge in modeling such genetic impact is the high-dimensionality and noise associated with GWAS data. Therefore, we deployed a generative model on the genetic data, assuming that there exist a few distinct genomic makeups for a particular disease, which manifest at different degrees in each individual sample.

To parameterize the inference models $q_\phi$ and generative models $p_\theta$ in the first two terms of \cref{eq:elbo_final}, we propose a deep-learning algorithm based on a variational auto-encoder (VAE).
For better interpretability and inference of latent states, we select a one-layer neural network as the VAE decoder. Specifically, the decoder is parameterized by a weight matrix $\mathbf{S} \in \mathbb{R}^{K\times d_g}$, where $K$ is the number of genetic groups, and $d_g$ is the dimension of genetic features.
Intuitively, the weight matrix $\mb S$ decodes the distribution of genetic features for each cluster, and $\mb V$ represents the assignments of each sample to a cluster $k \in {1,2,...,K}$. 
It's important to note that we don't require $\mathbf{V}$ to be one-hot encoded in this context. 
The cluster assignment is done in a soft manner, allowing a single patient to be assigned to multiple clusters based on probability. 
The input of this module will be the genetic data $\mb G$. The clustering assignments $\mb V$ and $\mb S$ are learned jointly with the transition functions. More details are provided in \cref{al:pdpm}.

\begin{align}
        \label{eq:vae}
        \boldsymbol{\mu}_v, \boldsymbol{\sigma}_v &= Encoder(\mb G)\nonumber\\
        \mb V &\sim \mathcal{N}(\boldsymbol{\mu}_v, \boldsymbol{\sigma}_v)\\
        \hat{\mb G} &= \mb V\mb S + b 
\end{align}

\subsection{Genetics Driven State Transition} 
In this subsetcion, we discuss how to design the genetics driven state transition, corresponding to $q_\phi(\mb Z_t|\mb Y_t, \mb Z_{t-1})$ and $p_\theta(\mb Z_t|\mb Z_{t-1}, \mb V)$.

First, we use a GRU base inference network to capture the variational
distribution of  $q_\phi(\mb Z_t|\mb Y_t, \mb Z_{t-1})$ and sample the posterior. Specifically,
\begin{align}
    \mb h_{\mb z, t} = GRU([\mb X_t, \mb U_t, G])\nonumber\\
    \Tilde{\boldsymbol{\mu}}_{\mb z, t}, \Tilde{\boldsymbol{\sigma}}_{\mb z, t} = C(\mb h_{\mb z, t}, [\mb X_t, \mb U_t, G])\nonumber\\
    \Tilde{\mb Z}_t \sim \mathcal{N}(\Tilde{\boldsymbol{\mu}}_{\mb z, t}, \Tilde{\boldsymbol{\sigma}}_{\mb z, t})
    \label{eq:inf_model}
\end{align}
where $C$ is combiner function adding a skip connection to inference network, which is widely used in structural inference based methods \cite{krishnan2017structured, alaa2019attentive, hussain2021neural}.

For $p_\theta(\mb Z_t|\mb Z_{t-1}, \mb V)$, we designed a genetics-driven state transition module.
Given the assumptions of progression being conditioned on these genetic makeups and treatments, we use a set of transition functions $f_i$ to model the genetics-driven disease progressions. 
To ensure that transition functions are conditioned on the genetic makeups, we associate each transition function with one genetic cluster as follows:
\begin{align}
    \mb F_t = [f_1(\Tilde{\mb Z}_{t-1}, \mb U_{t}, \mb S_1), \dots,f_K(\Tilde{\mb Z}_{t-1}, \mb U_{t},  \mb S_K)]\label{eq:trans1}
\end{align}

As shown in \cref{eq:trans1}, each transition function takes the corresponding genetic makeups $\mb S_k$ as input, together with previous latent state $\mb Z_{t-1}$ and treatment $\mb U_{t}$. 
We then use an attention mechanism to aggregate genetics-driven disease progressions. The weights of the attention are determined by the softmax of $\mb V$, which is jointly inferred with genetic makeups.
\begin{align}
    \boldsymbol{\mu_{\mb z, t}}, \boldsymbol{\sigma_{\mb z, t}}  =softmax( \mb V)\mb F_t
    \label{eq:trans_model}
\end{align}

For the rest of the model, we used 2-layers neural networks with ReLU activation for generative model $p_\theta(\mb X_t| \mb Z_t)$, i.e., $\boldsymbol{\mu_{x,t}}, \boldsymbol{\sigma_{x,t}} = NN(\Tilde{\mb Z}_{t})$.
We provide additional details for the PerDPM model in \cref{al:pdpm}. The combiner function is presented in \cref{al:Combiner}.

\subsection{Objective Function}
Using aforementioned methods, the objective function can be written as follow
following \cref{eq:elbo_final}: 
\begin{align}
    \mathcal{L} =  MSE_{VAE} + KL_{VAE} + NLL + KL_z
    \label{eq:EBLO_loss}
\end{align}
where
\begin{align}
    &MSE_{VAE} = MSE(\mb G, \hat{\mb G})\nonumber\\
    &KL_{VAE} = KL(\mathcal{N}(0, 1)||\mathcal{N}(\mu_{v}, \sigma_{v}))\nonumber\\
    &NLL = - \text{log-likelihood}(\mb X, \boldsymbol{\mu_{x}}, \boldsymbol{\sigma_{x}} )\nonumber\\
    &KL_z = KL(\mathcal{N}(\Tilde{\boldsymbol{\mu}}_{\mb z, t}, \Tilde{\boldsymbol{\sigma}}_{\mb z, t})||\mathcal{N}(\boldsymbol{\mu_{\mb z, t}}, \boldsymbol{\sigma_{\mb z, t}}))
\end{align}





\begin{algorithm}
	\caption{Learning PerDPM} 
 \label{al:pdpm}
	\begin{algorithmic}[1]
        \State Input: $\mb X$ (Observations), $\mb U$ (Treatment),  $\mb G$ (Genetics)
        \State Output: 
        $\mb Z$ (State Variables)
        \State Genetic makeups Inference: $q_\phi(\mb V|\mb G)$, $p_\theta(\mb G|\mb V)$
        \State \quad Sample $\mb V \sim q_\phi(\mb V|\mb G)$ \hspace*{0pt}\hfill\cref{eq:vae}
        \State \quad Estimate $\mb S$ from $p_\theta(\mb G|\mb V)$ 
        \State Inference Network: $q_\phi(\mb Z_t|\mb Y_{t-1}, \mb Z_{t-1})$, $p_\theta(\mb Z_t|\mb Z_{t-1}, \mb V)$
        \State \quad Sample $\Tilde{\boldsymbol{\mu}}_{\mb z, t}, \Tilde{\boldsymbol{\sigma}}_{\mb z, t} \sim q_\phi(\mb Z_t|\mb Y_{t-1}, \mb Z_{t-1})$ \hspace*{0pt}\hfill\cref{eq:inf_model}
        \State \quad Sample $\boldsymbol{\mu_{\mb z, t}}, \boldsymbol{\sigma_{\mb z, t}} \sim p_\theta(\mb Z_t|\mb Z_{t-1}, \mb V)$  \hspace*{0pt}\hfill\cref{eq:trans_model}
        \State Generative Network: $p(\mb X|\mb Z)$
        \State \quad Sample $\boldsymbol{\mu}_{xt}, \boldsymbol{\sigma}_{xt} \sim p(\mb X_t|\mb Z_t)$
        \State Loss
        \State \quad Estimate Objective Function $\mathcal{L}$  \hspace*{0pt}\hfill\cref{eq:EBLO_loss}
        \State Estimate the gradient $\nabla_\phi \mathcal{L}, \nabla_\theta \mathcal{L}$
        \State Update $\phi, \theta$ jointly
	\end{algorithmic} 
\end{algorithm}

\begin{algorithm}
	\caption{ Combiner Function: $C$} 
	\begin{algorithmic}[1]
        \State Input: $\mb h_{1}$, $\mb h_{2}$
        \State Output $\mu$, $\sigma$
        \State Compute $\mu$ and $\sigma$
        \State \quad $mu1, sig1  = NN(h_{1}), Softplus(NN(h_{1}))$ 
        \State \quad $mu2, sig2  = NN(h_{2}), Softplus(NN(h_{2}))$
        \State \quad $sigmasq = \frac{sig1 + sig2}{sig1^2 + sig2^2}$
        \State \quad $\mu = (mu1/sigsq1 + mu2/sigsq2)*sigmasq$
        \State \quad $\sigma = \sqrt{sigmasq}$
	\end{algorithmic} 
 \label{al:Combiner}
\end{algorithm}

\section{Experiment on Synthetic Dataset}

\subsection{Synthetic Dataset}
\label{sec:syn_dat}
A synthetic dataset is designed to simulate the progression of chronic disease, driven by genetic makeups. 
The generation process consists of the generation of genetic information, treatment, disease states and clinical observations.

\subsubsection{Generation of Genetic Markers}
As shown in \cref{al:synthetic_dat} lines 1-10,
we first generated genetic markers for each patient as their inherent characteristics. 
Specifically, we first created genetic clusters and the corresponding genetic markers distributions $\mb S$. 
Then, a cluster assignment variable $\mb V$ is created for each patient. 
The final genetic features were then derived by computing the average of the genetic marker distributions $\mb S$, with weights determined by the cluster assignment variables $\mb V$.

\subsubsection{Generation of Treatments}
To generate binary treatment variables, we randomly sample two time indices, $t_{\text{st}}$ and $t_{\text{end}}$, indicating the start and ending points of treatments. Then, we set the corresponding entries to 1 in the treatment sequence, as shown in lines 11-14 in \cref{al:synthetic_dat}.

\subsubsection{Generation of Disease States}
The disease states are latent variables describing the disease severity. The transition between different states depends,in reality, on multiple complex factors. In this synthetic dataset, we simplify the transitions model and make the disease progression driven by genetic markers.
In lines 15-20 within \cref{al:synthetic_dat}, we created a transition function $f_k$ for each genetic cluster $k$. The state is computed by aggregating all $K$ transitions using the cluster assignment $\mb V$, followed by a softmax function.
The initial disease states are sampled from a multivariate normal distribution, with individual-specific mean values (not genetic cluster-specific means).


\subsubsection{Generation of Clinical Observations}
The clinical observations are generated from the disease state through an emission model, parameterized by a linear model, as demonstrated in lines 22-23 in \cref{al:synthetic_dat}.

\begin{algorithm}[h]
	\caption{Genetic Synthetic Dataset} 
 \label{al:synthetic_dat}
	\begin{algorithmic}[1]
        \State Genetic Information Generation - Output: Genetic markers $\mb G$
        \State \quad \# Create mean and covariance for each genetic cluster
        \State \quad $\boldsymbol{\mu}_{g} = U(-5, 5) \in \mathbb{R}^{K\times d_g}$
        \State \quad $\boldsymbol{\sigma}_{g} = [\mb I_g * U_k(0, 1)]_K  \in \mathbb{R}^{K\times  d_g \times d_g}$
        \State \quad \# Create genetic features for each genetic cluster
        \State \quad $\mb S \sim \mathcal{N}\left(\boldsymbol{\mu}_{g}, \boldsymbol{\sigma}_{g}\right) \in \mathbb{R}^{K\times N \times d_g}$
        \State \quad \# Create cluster assignment for each data
        \State \quad $\mb V \sim Cat(\mathcal{N}(0, 1))\in \mathbb{R}^{K\times N}$
        \State \quad \# Create genetic information using weighted average
        \State \quad $\mb G = \mb V * \mb S\in \mathbb{R}^{N\times d_g}$
        \State Treatment Generation - Output: Treatment sequence $\mb U$
        \State \quad $\mb U = \boldsymbol{0} \in \mathbb{R}^{N\times T\times d_u}$
        \State \quad $t_{st}, t_{end} \sim U(0, Tmax) $
        \State \quad $\mb U[t_{st}: t_{end}] = 1$
        
        \State Generate initial disease state - Output: $\mb Z_{init} $
        \State \quad $\boldsymbol{\mu}_{init} = \mathcal{N}(\boldsymbol{0}, \boldsymbol{1})$
        \State \quad $\mb Z_{init}\sim \mathcal{N}(\boldsymbol{\mu}_{init}, \boldsymbol{1})$
        
        \State Generate disease states - Output: $\mb Z $
        \State \quad $\mb h_{t} = [f_{1}( [\mb Z_{t-1}, \mb U_{t-1}] ), f_{K}( [\mb Z_{t-1}, \mb U_{t-1}] )]  \in \mathbb{R}^{N\times K\times d_z}$ 
        \State \quad $\mb Z_{t} = softmax(\mb V\mb h_{t})  \in \mathbb{R}^{N\times d_z}$
        \State Generate clinical observations - Output:$\mb X$
        \State \quad $\mb X_t \sim \mathcal{N}(\mb w_{x} \mb Z_t+ b_{x}, \mb I)$
	\end{algorithmic} 
\end{algorithm}
\subsection{Comparison Methods and Evaluation Metrics}
We selected three state-of-the-art methods as comparison methods. All of these method are based on state space methods.

\textit{Deep Markov Models (DMM) \cite{krishnan2017structured}}: DNN is an algorithm designed to learn non-linear state space models. Following the structure of hidden Markov models (HMM), DMM employs multi-layer perceptrons (MLPs) to model the emission and transition distributions . The representational power of deep neural networks enables DMM to model high-dimensional data while preserving the Markovian properties of HMMs.

\textit{Attentive State-Space Model (ASSM) \cite{alaa2019attentive}}: ASSM learns discrete representations for disease trajectories. ASSM introduces the attention mechanism to learn the dependence of future disease states on past medical history. In ASSM, state transitions are approximated by a weighted average of the transition probability from all previous states. i.e., $q_\phi(\mb Z_t|\mb Z_{t-1}) = \sum_{i = 1}^{t-1}\alpha_{i}\mb P(\mb Z_i, \mb Z_t)$,
where $\alpha_{i}$ is a learnable attention weight, and $\mb P$ is the predefined transition matrix obtained from observations $\mb X$ using a Gaussian Mixture model. 

\textit{Intervention Effect Functions (IEF)  \cite{hussain2021neural}}: IEF-based disease progression model is a deep state space model designed for learning the effect of drug combinations, i.e., pharmacodynamic. IEF proposes an attention based neural network architecture to learn the pharmacodynamic, i.e., how treatments affect disease states.
Although IEF is not tailored for genetics-driven progression models, it has the capacity of incorporating individual-specific covariates. Thus, in the experiments, we test two variants of IEF: one without using genetic information (named IEF) and one using genetic information as static covariates (named IEF w/ G).

For the evaluation metrics, we report the test negative log likelihood (NLL) of clinical observations on synthetic dataset. In the real-world dataset, the clinical observations are binary variables, and we reported the Cross-Entropy (CE) instead.
We also reported the Pearson's chi-square statistic (Chi2) between the predicted states and the true states, calculated using a contingency table. 
The Chi2 score measure the discrepancies between the predicted results and the expected frequencies, which can be seen as the random guess in our case. 
Thus, a higher Chi2 score is better. For both NLL and CE score, lower values are better.

\subsection{Results on Synthetic Dataset}
Utilizing the data generation method introduced in \cref{sec:syn_dat}, we created 5 different datasets using with various settings. 
The number of disease states and the number of genetic clusters are both set to 5 for all these five datasets. 
For each dataset, we performed 10 random train-test splits and conducted evaluations using comparison methods.
The results for the synthetic dataset 1 are reported in \cref{tab:result_syn_data} (10 experiments). 
Additionally, we report the results for all the synthetic dataset in \cref{tab:result_syn_data_all} (50 experiments) in the Appendix. 
It's worth noting that the comparison method ASSM does not maximize the likelihood of clinical observations directly and doesn't provide the estimated mean and variance for the covariates. Therefore, we didn't report the negative log likelihood (NLL) for ASSM.

As shown in \cref{tab:result_syn_data}, DMM shows the worst performance compared to all other methods. Similar to a hidden Markov model, DMM assumes that all patients follow the same transition pattern, as the transitions between latent states are modeled by the same neural networks. When the latent dynamic of state transition varies according to patient's inherited characters, the assumption of DMM fails. Modeling the transition uniformly prohibits DMM to capture genetic driven disease progression in our synthetic dataset.

Compared to DMM, ASSM and IEF incorporate attention mechanisms into their transition models, enabling them to capture the variations in transition patterns. In \cref{tab:result_syn_data}, ASSM and IEF show better scores than DMM in both NLL and Chi2. 
Considering the similarity in the generative models of IEF and DMM, the lower NLL for IEF suggests it learned a better representation of latent states $\mb Z$ by using attention mechanisms.
Also, higher Chi2 scores suggest the inferred latent states from IEF and ASSM are closer to the true states than those inferred from DMM.

When incorporating genetic information, the IEF w/ G can be seen as the personalized model. 
The difference in performance between IEF w/ G and IEF indicates that using genetic information is helpful for personalized disease modeling. 
However, the performance gaps between IEF w/G and IEF become smaller as the training set size decreases. 
This suggests that the simple concatenation of genetic information and clinical observation is insufficient to learn the genetic clusters and the disease progression pattern it drives, especially in the case when the genetic information is noisy.

\begin{table*}[]
    \centering
        \caption{Result on a synthetic dataset across 10 different runs. Both negative log likelihood (lower values are better) and Chi-square statistic (higher values are better) along with standard deviation are calculated to evaluate our models. We reported different results as the training set size varies from 600 to 3000.}
    \begin{tabular}{|c|c|c|c|c|c|c|}
    \hline
        Methods & ASSM &DMM&IEF&IEF w/ G&PerDPM& Trainin Set Size\\
        \hline
       NLL  &$-$&$52.29\pm4.30$& $11.92\pm1.03$&$11.96\pm0.18$&\boldsymbol{$8.65\pm0.57$}&\multirow{2}{*}{600}\\
       \cline{0-5}
       Chi2 & $948.17\pm877.03$ & $661.52\pm332.38$& $1671.97\pm366.55$&$1662.79\pm145.07$&\boldsymbol{$2236.43\pm266.41$}&\\
       \hline \hline
              NLL  &$-$&$96.87\pm1.35$& $6.17\pm0.51$&$6.37\pm1.25$&\boldsymbol{$1.16\pm2.15$}&\multirow{2}{*}{1500}\\
       \cline{0-5}
       Chi2 & $3810.95\pm3018.58$ & $218.42\pm60.68$& $5024.76\pm330.52$&$5603.98\pm361.83$&\boldsymbol{$7299.05\pm1004.70$}&\\
       \hline \hline
        NLL  &$-$&$192.05\pm2.59$& $3.33\pm1.78$&$1.38\pm2.00$&\boldsymbol{$-1.72\pm2.46$}&\multirow{2}{*}{3000}\\
       \cline{0-5}
       Chi2 & $6512.63\pm5555.98$ & $1064.73\pm914.85$& $9922.56\pm610.88$&$10934.06\pm736.13$&\boldsymbol{$15601.18\pm2327.51$}&\\
       \hline

    \end{tabular}

    \label{tab:result_syn_data}
\end{table*}

\begin{figure}
    \centering
    \includegraphics[width=\linewidth]{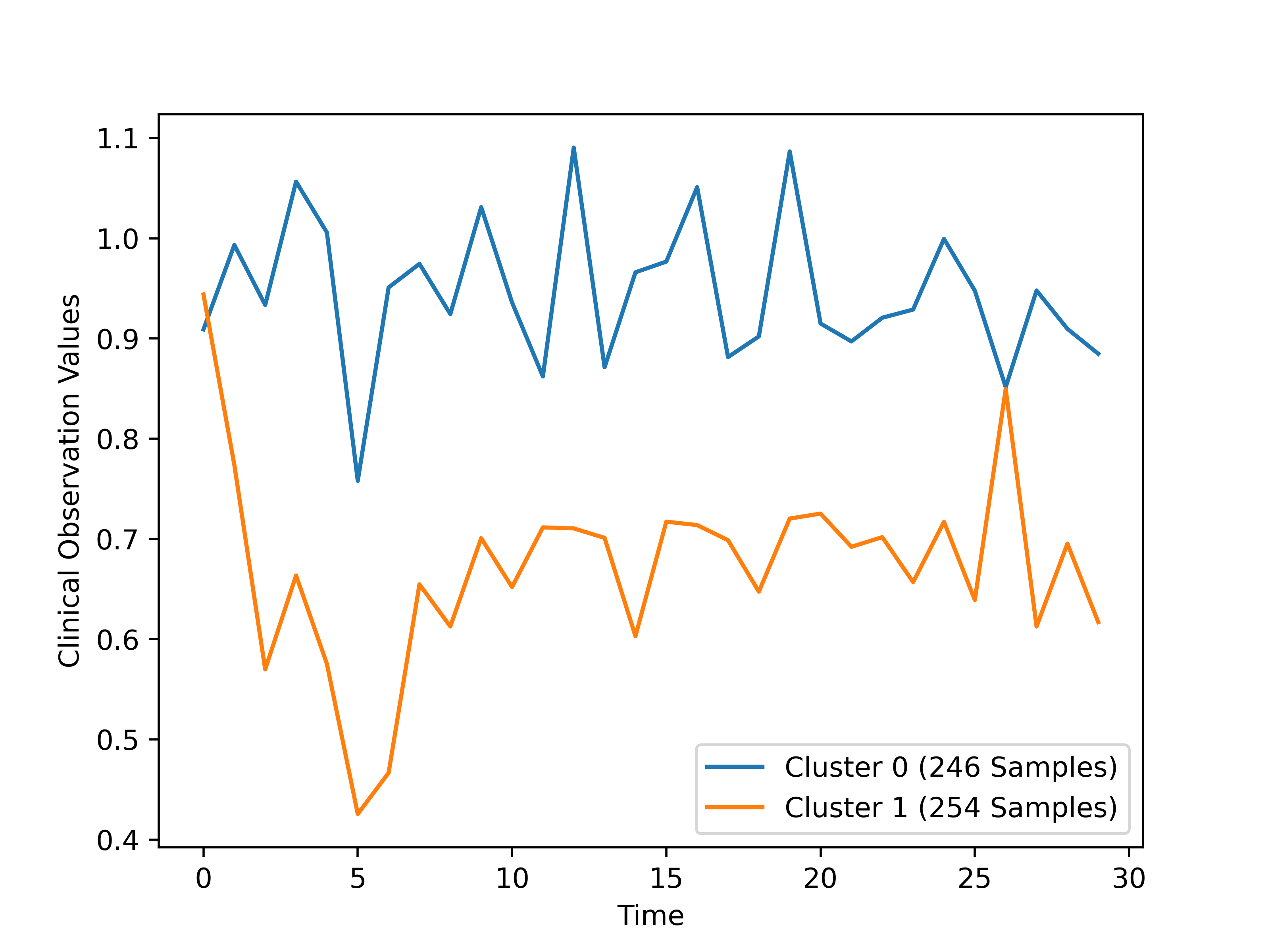}
    \caption{Analysis on synthetic datasets: The plot shows the mean values of clinical observations within different sample groups. The sample groups are discovered using the cluster variable $\mb V$, which is inferred by our proposed model. The x-axis represents the time steps, and the y-axis represents the clinical observation values. The plot demonstrates the latent variable $\mb V$ in our proposed model has the capacity to discriminate between patient groups with different disease progression types.}
    \label{fig:syn_genetic}
\end{figure}

Our model outperforms all other methods in both metrics, and the improvement becomes more significant as the sample size increase.
To further assess the effectiveness of our model, we conducted additional analysis on the discovered genetic groups to assess how well they can segregate the disease progression patterns. 
We used KMeans to learn two genetic clusters using the representations of the inferred clusters variables $\mb V$. 
Then, we plotted the mean values of one dimension of observation, i.e., $\mb X^{(0)}$, along the timeline for each cluster in \cref{fig:syn_genetic}. Clearly, samples from these two clusters have very different observation values, although these two clusters have similar initial values. This demonstrates that our designed genetics-driven state transition can effectively identify different transition patterns governed by diverse genetic makeups for different genetics clusters.



\section{Results on modeling progression of Chronic Kidney Disease}
\subsection{Real-World Chronic Kidney Disease Dataset}
Chronic kidney disease (CKD) is a condition where the kidneys are damaged and progressively lose their ability to filter blood. It is estimated that $800$ millions or $10\%$ of the world population have CKD \cite{kovesdy2022} and $37$ millions in the USA alone, it being one of the leading causes of death while $90\%$ of adults with CKD and $40\%$ of adults with severe CKD do not know that they already have the disease \cite{cdc2021}. 
CKD is primarily defined in Clinical Practice Guidelines in terms of kidney function \cite{nationalkidneyfoundationDOQIClinicalPractice2002} and CKD patients progress over five CKD stages, often slowly and heterogenously \cite{abeysekera2021heterogeneity}, from mild kidney damage to End Stage Kidney Disease (ESKD) or kidney failure, defined as either the initiation of dialysis or kidney transplant. Later stages of CKD are defined based on lower levels of creatinine-based estimated glomerular filtration rate (eGFR below $60$ ml min $1.73$m$^{-2}$) hence capturing an heterogeneous set of kidney disorders. However, eGFR has been used in a set of genomic studies as a trait for finding common variants associated with kidney disease. GWAS can explain up to $20\%$ of an estimated $54\%$ heritability in this CKD-associated trait \cite{wuttke2019} and have helped establish genome-wide polygenic scores (GPS) across ancestries for discriminating moderate-to-advanced CKD from population controls \cite{khan2022}.

We used a large-scale real world dataset containing approximately 250,000 samples from the UK, collected as UKBioBank repository \cite{sudlow2015uk} to validate our proposed GWA-PerDPM model. 
We extracted clinical features such as clinical classification system (CCS) codes for measuring co-morbidities and therapeutic drug classes for measuring the treatments in the context of modeling chronic kidney disease (CKD). We also curated approximately 300 SNPs from the genome-wide-association (GWA) using the common quality control protocols. Furthermore, we computed the true stages of CKD based on the eGFR measurements that were available in the dataset.  

\subsection{Result of Chronic Kidney Disease Progression Modeling}
We reported the results on real-world chronic kidney disease dataset in \cref{tab:result_ckd_data}. 
In our CKD dataset, clinical observations consist of binary CCS codes, which makes ASSM infeasible for use since it employs conditional density estimation. Therefore, In this section, we only compare our proposed method with IEF since it is easy to modify to handle binary clinical observations. In both IEF and our method, we adjusted the ELBO by adopting cross entropy to accommodate the binary nature of the clinical observations, and we reported cross entropy (CE) instead of negative log-likelihood (NLL) in the table to measure how well the generative models fit the observations.

As shown in \cref{tab:result_ckd_data}, the results are consistent with those obtained from the synthetic dataset, demonstrating that our methods outperform other methods in both metrics. 
it's worth nothing that the score gap between PerDPM and IEF w/ G becomes larger in the CKD progression modeling. One reason could be the increased noise in clinical observations (we use the binary diagnosis code as clinical observation).
Also, genetic information shows less correlation with disease progression, considering the fact that GWAS can only explain up to 20\% of an estimated 54\% heritability in this CKD-associated trait. This emphasizes the necessity of modeling genetic driven dynamic explicitly.

\begin{figure}[h]
    \centering
    \includegraphics[width = 0.8\linewidth]{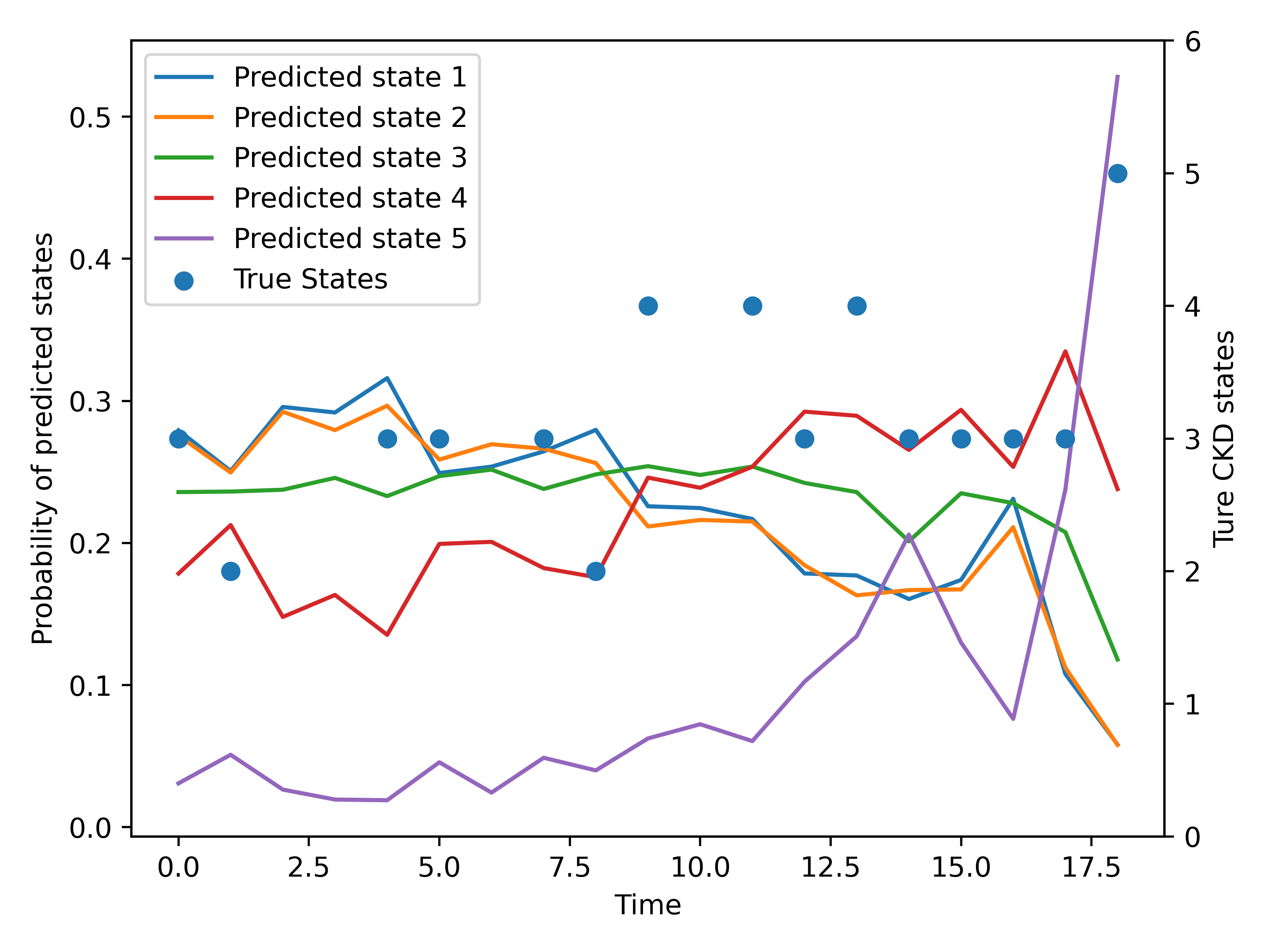}
    \caption{
    Analysis on Real-World Datasets: The plot illustrates the predicted CKD states (lines) versus the true CKD states (dots) for one patient along the timeline. Each line represents the probability of the predicted state, with the corresponding values displayed on the left-hand side of the Y-axis. Each dot represents the true CKD states graded by eGFR score, with the discretized state number positioned on the left-hand side of the Y-axis.
    }
    \label{fig:vis_ckd_states}
\end{figure}

In \cref{fig:vis_ckd_states}, we visualized the inferred hidden states from our method for one patient, along with the true CKD states. 
In the figure, we plot the probability of each inferred state as color-coded lines, with the corresponding values plotted on the left-hand side of the y-axis. 
For example, at time 0, the predicted probability of the patient being in CKD state 2 is about 0.28 since the orange line has value 0.28 at $t=0$.
The true CKD states are represented by the blue dots at the date when the eGFR is measured, which is the indicator of CKD severity.
These CKD states vary among $[1,2,3,4,5]$, reflecting the degrees of severity, with corresponding discrete state values shown on the right-hand side of the y-axis.
For example, the patient was diagnosed with CKD stage 3 at time 0, as the first blue dot reads number 3 on the right-hand side of the y-axis. 
As shown in \cref{fig:vis_ckd_states}, as the disease progress, the probability of predicted state 1, 2, and 3 gradually decrease, while the probabilities of predicted state 4, 5 gradually increase. The dynamic of the inferred states is consistent with the patient's true CKD progression.
If we choose the inferred state based on the highest predicted probability, the inferred states align well with the true states represented by the blue dots.

\begin{figure}[h]
    \centering
    \includegraphics[width = 0.85\linewidth]{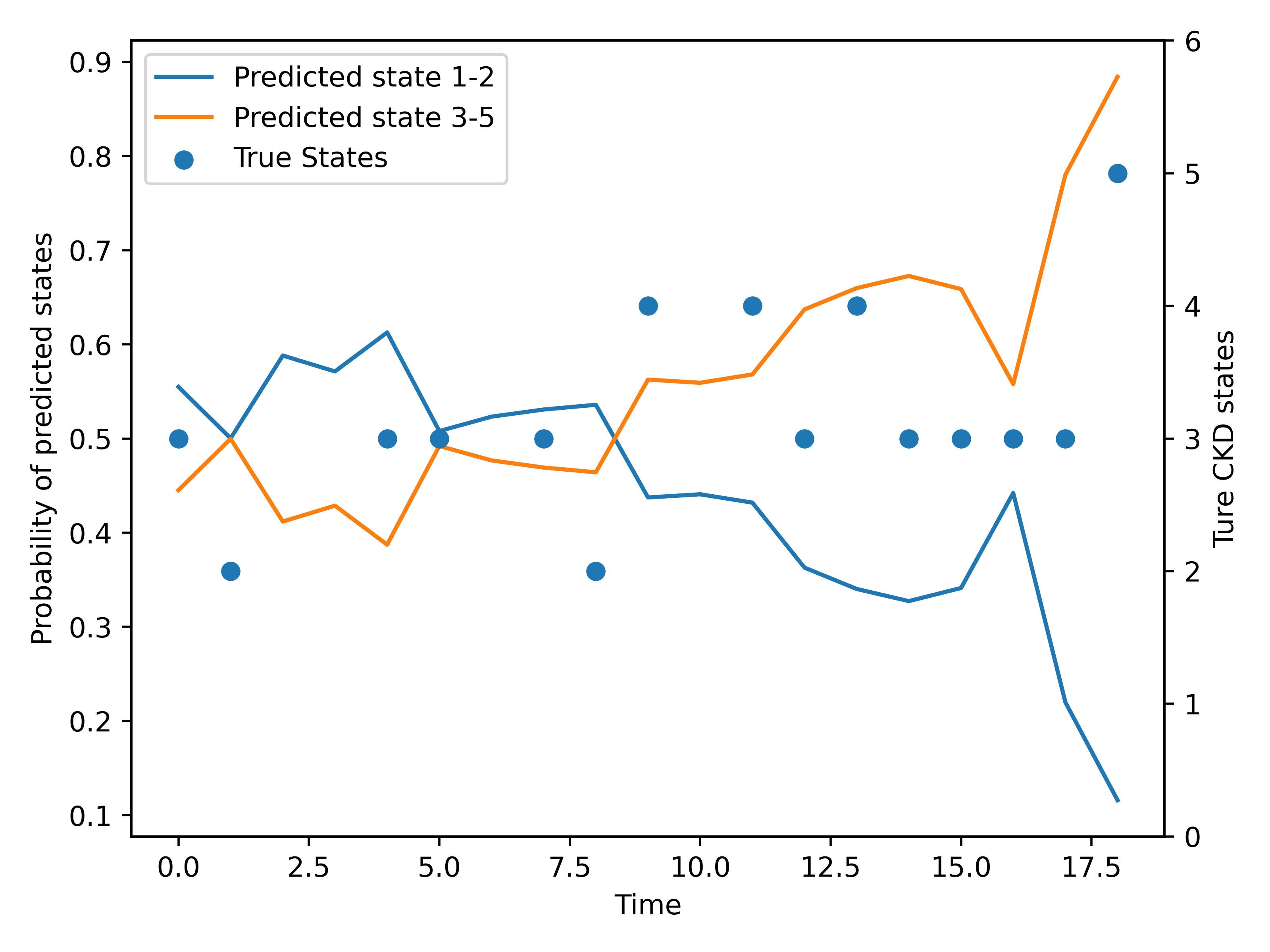}
    \caption{Analysis on Real-World Datasets: The plot displays the aggregated predicted CKD states (lines) versus the true CKD states (dots) for one patient along the timeline. The orange line represents the risk of having CKD, while the blue line represents the risk of not having CKD. The blue dots correspond to the true CKD stages, with a higher stage number indicating a more severe CKD disease.}
    \label{fig:vis_ckd_states2}
\end{figure}

In chronic kidney disease, CKD stage 1 and 2 correspond to mild conditions. People are usually considered as not having CKD in the absence of markers of kidney damage when they are in state 1 and 2. 
Following this criterion and for better visualization, we further aggregate the probability of states 1 and 2, as well as the probability of states 3-5 from our model, to binarize the predicted progression into CKD and no CKD. 
The visualization is shown in \cref{fig:vis_ckd_states2}. 
The probability of predicted states also aligns well with the true states represented by the blue dots.

\begin{figure}[h]
    \centering
    \includegraphics[width = 0.85\linewidth]{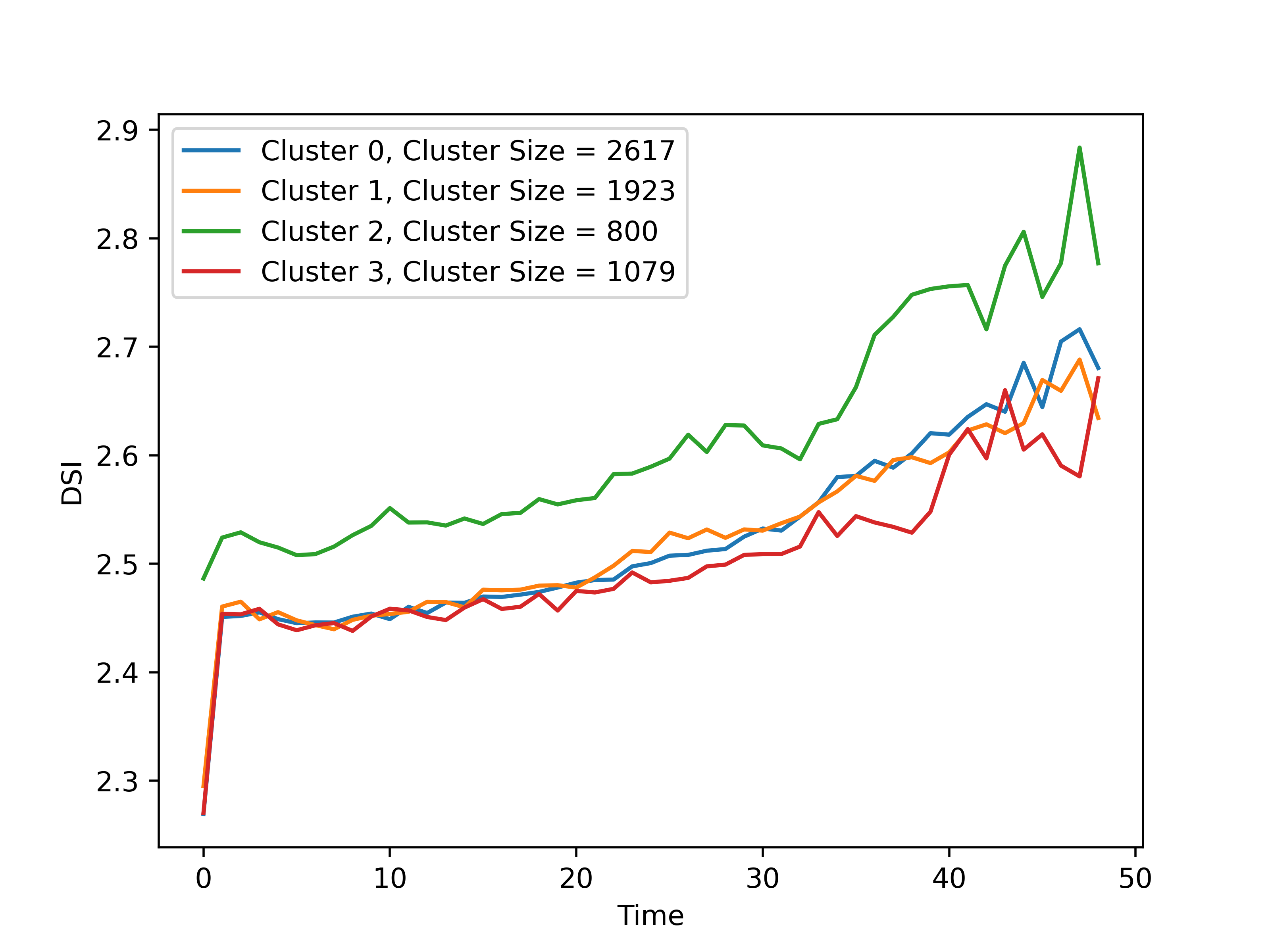}
    \caption{Analysis on Real-World Datasets: The plot shows the average disease severity index (DSI) in each predicted genetic cluster. Each line shows the change in mean DSI as disease progress. The DSI in the predicted cluster 2 is significantly higher than in other clusters, suggesting our method learned meaningful clusters from genetic markers.}
    \label{fig:vis_dsi_mean}
\end{figure}

To further analyse if the proposed model learned genetics driven progression, we plot the disease severity index (DSI) per each time steps for each genetic group in \cref{fig:vis_dsi_mean}. DSI is defined as the expected disease states by the following equation: $DSI = \sum_{k}\{z_{ijk} * p(z_{ijk})\}$, for $i$-th sample and $j$-th time point and each possible discrete states $k \in \{1,2,...,K\}$. 
In \cref{fig:vis_dsi_mean}, each line shows the change in mean DSI as disease progress in each genetic cluster. 
As shown in the figure, genetic cluster 2 exhibits different behavior than the other three clusters. 
Patients in cluster group 2 have higher CKD risk at their index day (first visit) and are more likely to progress into more severe stages. 
On the other hand, all other clusters started at same DSI, but patients in cluster 3 tend to have a lower DSI at later stages. 
This shows the presence of genetic impacts on the progression patterns in CKD and the ability of our model to capture them. 

\begin{table}[h]
    \centering
        \caption{Result on realworld chronic kidney disease dataset}
    \resizebox{\columnwidth}{!}{
    \begin{tabular}{|c|c|c|c|c|}
    \hline
        Methods &  IEF&IEF w/ G&PerDPM\\
        \hline
       CE  & $770.89\pm 5.32$&$520.00\pm 4.80$&$449.81\pm 2.27$\\
       \hline
       Chi2  & $41.64 \pm 7.13$&$135.95\pm38.65$&$1538.16\pm 272.27$\\
       \hline
    \end{tabular}
    }
    \label{tab:result_ckd_data}
\end{table}

\section{Conclusion}

In this paper, we developed a personalized disease progression model based on the heterogeneous genetic makeups, clinical observations, and treatments of individual patients. We jointly inferred the latent disease states and genetics-driven disease progressions using the proposed genetic makeups inference module and genetics-driven state transition module.
We demonstrated improvements over state-of-the-art methods using both synthetic data and a large-scale real-world dataset from the UK BioBank cohort of Chronic Kidney Disease.
Our analysis shows that the proposed model is generic enough to discover diverse genetic profiles and associated disease progression patterns. Therefore, it can be deployed in the future on a wide variety of healthcare datasets for deriving useful clinical insights.

\nocite{*}
\bibliography{ref}
\bibliographystyle{ACM-Reference-Format}

\clearpage
\newpage
\appendix
\section{Derivation of the Evidence Lower Bound}\label{sec:proof_of_elbo}
We using the evidence lower bound to approximate the likelihood in \cref{eq:likelihood_overall}. For simplicity, let $\mb Y = [\mb X, \mb G, \mb U]$, and $\mathcal D$ represent the dataset

\begin{align}
\label{eq:likelihood_and_elbo_appendix}
    \log p(\mb Y) \geq &\mathbb{E}_{\mb Y\sim\mathcal{D}, q_\phi(\mb Z, \mb V|\mb Y)}[\log p(\mb Y|\mb Z, \mb V)] \\
    &- 
     \mathbb{E}_{\mb Y\sim\mathcal{D}}[KL(q_\phi(\mb Z, \mb V|\mb Y)||p_\theta(\mb Z, \mb V))]
\end{align}
where $q_\phi$ and $p_\theta$ are learned posterior and prior distributions

Using \cref{eq:likelihood_overall} to expand this expression yields
\begin{align}
    ELBO =& \mathbb{E}_{\mb Y\sim\mathcal{D}, q_\phi(\mb Z, \mb V|\mb Y)}[\sum_t(\log p(\mb U_t) + \log p(\mb X_t| \mb Z_t) \nonumber\\
    &+ \log p(\mb G | \mb V))] \nonumber\\
    - &\mathbb{E}_{\mb Y\sim\mathcal{D}}[\sum_t KL(q_\phi(\mb Z_t, \mb V|\mb Y_t, \mb Z_{t-1})||p_\theta(\mb Z_t, \mb V))]
\end{align}
where term $\log p(\mb U_t)$ cannot be optimized. We can drop it and rewrite the equation as follow:
\begin{align}
\label{eq:elbo_simplified_appendix}
    ELBO =& \mathbb{E}_{\mb Y\sim\mathcal{D}, q_\phi(\mb Z, \mb V|\mb Y)}[\sum_t(\log p(\mb X_t| \mb Z_t) + \log p(\mb G | \mb V))] \nonumber\\
    &- 
     \mathbb{E}_{\mb Y\sim\mathcal{D}}[\sum_t KL(q_\phi(\mb Z_t, \mb V|\mb Y_t, \mb Z_{t-1})||p_\theta(\mb Z_t, \mb V))]
\end{align}

For the first term in \cref{eq:elbo_simplified_appendix}, we can break the expectation into two components as follow:
\begin{align}
\label{eq:elbo_first_term_decomp}
    &\mathbb{E}_{\mb Y\sim\mathcal{D}, q_\phi(\mb Z, \mb V|\mb Y)}[\sum_t(\log p(\mb X_t| \mb Z_t) + \log p(\mb G | \mb V))]\nonumber\\
     = &\mathbb{E}_{\mb Y\sim\mathcal{D}, q_\phi(\mb Z, \mb V|\mb Y)}[\sum_t\log p(\mb X_t| \mb Z_t)]\nonumber \\
     &+ T\mathbb{E}_{\mb Y\sim\mathcal{D}, q_\phi(\mb Z, \mb V|\mb Y)}[\log p(\mb G | \mb V)] 
\end{align}
The two components correspond to the reconstruction of observation $\mb X$ and genetic info $\mb G$, given the posterior of latent variables.

For the second KL divergence term in \cref{eq:elbo_simplified_appendix}, we have two distribution, posterior $q_\phi(\mb Z_t, \mb V|\mb Y_t, \mb Z_{t-1})$ and prior $p_\theta(\mb Z_t, \mb V)$.
When conditioned on the observational data, we assume that the posterior distribution is factorizable as follow:
\begin{align}
    q_\phi(\mb Z_t, \mb V|\mb X_t, \mb U_t, \mb Z_{t-1}, \mb G) = q_\phi(\mb Z_t|\mb X_t, \mb Z_{t-1}, \mb U_t, \mb V)q_\phi(\mb V|\mb G)
\end{align}
check if we need dv

Similarly,

\begin{align}
    p_\theta(\mb Z_t, \mb V) = p_\theta(\mb Z_t|\mb V)p_\theta(\mb V)
\end{align}

Thus, we have
\begin{align}
    &KL(q_\phi(\mb Z_t, \mb V|\mb Y_t, \mb Z_{t-1})||p_\theta(\mb Z_t, \mb V)) \nonumber\\
    =& KL( q_\phi(\mb Z_t|\mb X_t, \mb U_t, \mb Z_{t-1}, \mb V)q_\phi(\mb V|\mb G)|| p_\theta(\mb Z_t|\mb V)p_\theta(\mb V))\\
    = &\int q_\phi(\mb Z_t|\mb X_t, \mb U_t, \mb Z_{t-1}, \mb V)q_\phi(\mb V|\mb G)
    \log\frac{q_\phi(\mb Z_t|\mb X_t, \mb U_t, \mb Z_{t-1}, \mb V)q_\phi(\mb V|\mb G)}
    {p_\theta(\mb Z_t|\mb V)p_\theta(\mb V)} d\mb Z_t d\mb V\\
    =& \int q_\phi(\mb Z_t|\mb X_t, \mb U_t, \mb Z_{t-1}, \mb V)q_\phi(\mb V|\mb G)
    \log\frac{q_\phi(\mb Z_t|\mb X_t, \mb U_t, \mb Z_{t-1}, \mb V)}
    {p_\theta(\mb Z_t|\mb V)} d\mb Z_t d\mb V\nonumber\\
    &+ 
    \int q_\phi(\mb Z_t|\mb X_t, \mb U_t, \mb Z_{t-1}, \mb V)q_\phi(\mb V|\mb G)
    \log\frac{q_\phi(\mb V|\mb G)}
    {p_\theta(\mb V)}
    d\mb Z_t d\mb V\label{eq:KL1}
\end{align}
Note that 
\begin{align}
    &\int q_\phi(\mb Z_t|\mb X_t, \mb U_t, \mb Z_{t-1}, \mb V)q_\phi(\mb V|\mb G)
    \log\frac{q_\phi(\mb Z_t|\mb X_t, \mb U_t, \mb Z_{t-1}, \mb V)}
    {p_\theta(\mb Z_t|\mb V)} d\mb Z_t d\mb V  \nonumber\\
    =&\mathbb{E}_{q_\phi(\mb V|\mb G)}[KL(q_\phi(\mb Z_t|\mb X_t, \mb U_t, \mb Z_{t-1}, \mb V)||p_\theta(\mb Z_t|\mb V))]\label{eq:KL2}
\end{align}


and

\begin{align}
    &\int q_\phi(\mb Z_t|\mb X_t, \mb U_t, \mb Z_{t-1}, \mb V)q_\phi(\mb V|\mb G)
    \log\frac{q_\phi(\mb V|\mb G)}
    {p_\theta(\mb V)}
    d\mb Z_t d\mb V  \nonumber\\
    =&\int[\int q_\phi(\mb Z_t|\mb X_t, \mb U_t, \mb Z_{t-1}, \mb V)d\mb Z_t] q_\phi(\mb V|\mb G)
    \log\frac{q_\phi(\mb V|\mb G)}
    {p_\theta(\mb V)}
     d\mb V  \nonumber\\
    =&\int  q_\phi(\mb V|\mb G)
    \log\frac{q_\phi(\mb V|\mb G)}
    {p_\theta(\mb V)} d\mb V \nonumber\\
    =&KL(q_\phi(\mb V|\mb G)|| p_\theta(\mb V))\label{eq:KL3}
\end{align}

Combining \cref{eq:KL1}, \cref{eq:KL2}, and \cref{eq:KL3}, we have
\begin{align}
\label{eq:KL_final}
    &KL(q_\phi(\mb Z_t, \mb V|\mb Y_t, \mb Z_{t-1})||p_\theta(\mb Z_t, \mb V))\nonumber\\
    = &\mathbb{E}_{q_\phi(\mb V|\mb G)}[KL(q_\phi(\mb Z_t|\mb X_t, \mb U_t, \mb Z_{t-1}, \mb V)||p_\theta(\mb Z_t|\mb V))]\nonumber\\
    &+ KL(q_\phi(\mb V|\mb G)|| p_\theta(\mb V))
\end{align}

Combining \cref{eq:likelihood_and_elbo}, \cref{eq:elbo_first_term_decomp}, and \cref{eq:KL_final}, we have

\begin{align}
\label{eq:elbo_final_proof}
    \log p(\mb Y) \geq &\mathbb{E}_{\mb Y\sim\mathcal{D}, q_\phi(\mb Z, \mb V|\mb Y)}[\log p(\mb Y|\mb Z, \mb V)]\nonumber\\ 
    &- 
     \mathbb{E}_{\mb Y\sim\mathcal{D}}[KL(q_\phi(\mb Z, \mb V|\mb Y)||p_\theta(\mb Z, \mb V))]\nonumber\\
     = &\mathbb{E}_{\mb Y\sim\mathcal{D}, q_\phi(\mb Z, \mb V|\mb Y)}[\sum_t\log p(\mb X_t| \mb Z_t)] \nonumber\\
     &+ T\mathbb{E}_{\mb Y\sim\mathcal{D}, q_\phi(\mb Z, \mb V|\mb Y)}[\log p(\mb G | \mb V)]  \nonumber\\
     &- 
     \mathbb{E}_{\mb Y\sim\mathcal{D}}[KL(q_\phi(\mb Z, \mb V|\mb Y)||p_\theta(\mb Z, \mb V))]\nonumber\\
     = &\mathbb{E}_{\mb Y\sim\mathcal{D}, q_\phi(\mb Z, \mb V|\mb Y)}[\sum_t\log p(\mb X_t| \mb Z_t)] \nonumber\\
     &+ T\mathbb{E}_{\mb Y\sim\mathcal{D}, q_\phi(\mb Z, \mb V|\mb Y)}[\log p(\mb G | \mb V)]  \nonumber\\
     &-\mathbb{E}_{\mb Y\sim\mathcal{D}}[\sum_t\mathbb{E}_{q_\phi(\mb V|\mb G)}[\nonumber\\
     &KL(q_\phi(\mb Z_t|\mb X_t, \mb U_t, \mb Z_{t-1}, \mb V)||p_\theta(\mb Z_t|\mb V))]\nonumber\\
     &+ KL(q_\phi(\mb V|\mb G)|| p_\theta(\mb V))]\nonumber\\
     =&\underbrace{ T\mathbb{E}_{\mb Y\sim\mathcal{D}, q_\phi(\mb Z, \mb V|\mb Y)}[\log p(\mb G | \mb V)] }_{\text{Modelled by VAE}}\nonumber\\
    & - \underbrace{T\mathbb{E}_{\mb Y\sim\mathcal{D}}[KL(q_\phi(\mb V|\mb G)|| p_\theta(\mb V))] }_{\text{Modelled by VAE}}\nonumber\\
     &+\underbrace{\mathbb{E}_{\mb Y\sim\mathcal{D}, q_\phi(\mb Z, \mb V|\mb Y)}[\sum_t\log p(\mb X_t| \mb Z_t)] }_{\text{Modelled by State Space Model}}\nonumber\\
    &- \underbrace{\sum_t \mathbb{E}_{\mb Y\sim\mathcal{D}, q_\phi(\mb V|\mb G)}[KL( q_\phi(\mb Z_t|\mb X_t, \mb U_t, \mb Z_{t-1}, \mb V)|| p_\theta(\mb Z_t|\mb V))] }_{\text{Modelled by State Space Model}}
\end{align}

\section{Additional Experiments Results on Synthetic Dataset}

\begin{table*}[]
    \centering
        \caption{Result on 5 different synthetic datasets with 50 different runs in total. Both negtive log likelihood (lower is better) and Chi-square statistic (higher is better) standard deviation are calculated to evaluate our model}
    \begin{tabular}{|c|c|c|c|c|c|c|}
    \hline
        Methods & ASSM &DMM&IEF&IEF w/ G&PerDPM& Trainin Set Size\\
        \hline
       NLL  &$-$&$49.61\pm5.7$& $10.51\pm2.97$&$10.27\pm3.46$&\boldsymbol{$9.03\pm2.88$}&\multirow{2}{*}{600}\\
       \cline{0-5}
       Chi2 & $1439.23\pm1113.23$ & $639.15\pm462.07$& $1700.08\pm784.19$&$ 1811.00\pm749.17$&\boldsymbol{$2281.75\pm412.65$}&\\
       \hline \hline
              NLL  &$-$&$99.62\pm11.80$& $5.21\pm2.94$&$4.33\pm3.81$&\boldsymbol{$2.91\pm3.77$}&\multirow{2}{*}{1500}\\
       \cline{0-5}
       Chi2 & $4740.17\pm2919.49$ & $861.32\pm929.82$& $5696.28\pm1022.92$&$6695.59\pm1008.71$&\boldsymbol{$6966.07\pm1525.21$}&\\
       \hline \hline
        NLL  &$-$&$191.91\pm16.32$& $0.86\pm3.41$&$ -0.72\pm3.69$&\boldsymbol{$-0.50\pm3.73$}&\multirow{2}{*}{3000}\\
       \cline{0-5}
       Chi2 & $6954.37\pm5189.67$ & $1996.15\pm1573.52$& $13188.48\pm2313.02$&$ 14080.34\pm3033.52$&\boldsymbol{$14607.06\pm3021.49$}&\\
       \hline

    \end{tabular}

     \label{tab:result_syn_data_all}
\end{table*}

\end{document}